# Methodology and real-world applications of dynamic uncertain causality graph for clinical diagnosis with explainability and invariance

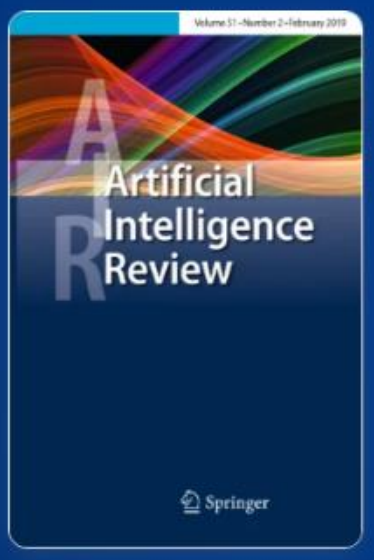

Zhan Zhang[1]· Qin Zhang[2]· Yang Jiao[3]· Lin Lu[4]· Lin Ma[5]· Aihua Liu[6]· Xiao Liu[7]·Juan Zhao[8]· Yajun Xue[9]· Bing Wei[10]· Mingxia Zhang[10]· Ru Gao[11]· Hong Zhao[10]·Jie Lu[12]· Fan Li[13]· Yang Zhang[14]· Yiming Wang[15]· Lei Zhang[16]· Fengwei Tian[17]·Jie Hu[18]· Xin Gou[19]

## Abstract

AI-aided clinical diagnosis is desired in medical care. Existing deep learning models lack explainability and mainly focus on image analysis. The recently developed Dynamic Uncertain Causality Graph (DUCG) approach is causality-driven, explainable, and invariant across different application scenarios, without problems of data collection, labeling, fitting, privacy, bias, generalization, high cost and high energy consumption. Through close collaboration between clinical experts and DUCG technicians, 46 DUCG models covering 54 chief complaints were constructed. Over 1,000 diseases can be diagnosed without triage. Before being applied in real-world, the 46 DUCG models were retrospectively verified by third-party hospitals. The verified diagnostic precisions were no less than 95%, in which the diagnostic precision for every disease including uncommon ones was no less than 80%. After verifications, the 46 DUCG models were applied in the real-world in China. Over one million real diagnosis cases have been performed, with only 17 incorrect diagnoses identified. Due to DUCG's transparency, the mistakes causing the incorrect diagnoses were found and corrected. The diagnostic abilities of the clinicians who applied DUCG frequently were improved significantly. Following the introduction to the earlier presented DUCG methodology, the recommendation algorithm for potential medical checks is presented and the key idea of DUCG is extracted.

**Key Words:** Diagnosis; Causality; Probabilistic Reasoning; Explainability; Counterfactual Inference

## 1. Introduction

Ref. (Rajpurkar et al. 2022) reviews many recent progresses in medical AI. It is seen that most medical AI models deal with image analysis. However, clinicians working at primary level need not only image analysis, but also others including comprehensive analysis of various symptoms, physical signs, laboratory and pathologic examinations, risk factors such as age, gender, post medical history, etc. In many cases (e.g. in village clinics), diagnoses are performed without medical images. Refs. (Liang et al. 2019) and (Wu et al. 2018) present two deep learning models for general disease diagnosis. However, the deep neural network (DNN) is a black-box approach without explainability. It is pointed out in Payrovnaziri and Chen (2020) that explainable AI (XAI) for medicine "is of vital importance to support the implementation of AI in clinical decision support systems" and "the new generation of AI systems have limited effectiveness due to the inability of humans to understand why an AI system makes particular decisions." In other words, a medical AI should have not only high diagnosis accuracy in the testing dataset and random clinical trials (RCTs), but also explainability to obtain trust from medical professionals, including to explain what and how medical knowledge is represented, how a diagnosis is inferred, and what is updated by adding more training data and what is the influence of the update, or briefly, "how the algorithm reaches its final decisions" (Payrovnaziri and Chen (2020)). However, "XAI evaluation in medicine has not been adequately and formally practiced" (Payrovnaziri and Chen (2020)). Ref. (Das and Rad 2020) presents similar concerns. Finding features does not have significant help to make DNN explainable. For example, local interpretable model-agnostic explanation (LIME) (Ribeiro et al. 2016) and Shapley additive explanation (SHAP) (Lundberg and Lee 2017) are two post hoc explanation methods. They can find which features contribute more to the diagnostic result according to certain statistical calculations. However, such post hoc explanations cannot internally explain to medical professionals why DNN reaches the diagnostic results instead of other results. Similarly, knowledge graph (KG) cannot explain why DNN reaches its diagnostic results, because KG is also external to DNN and its explanation is post hoc.

On the other hand, what clinicians most need is the correct diagnoses for uncommon diseases, not only for common diseases, because common diseases can usually be diagnosed by clinicians. However, DNN is trained with data. It is likely that the dominant data (the common disease case records) are well fitted but not the rare data (the uncommon disease case records), resulting in the lower accuracy to diagnose uncommon diseases, while the total diagnostic precision of DNN can still be high. That is, once the diagnostic precision for common diseases is high, the total diagnostic precision can be high, even though the diagnoses for uncommon diseases, which are really needed, are all incorrect, because the less but common diseases are dominant in the testing dataset and RCTs.

Yang Jiao, Lin Lu, Lin Ma, Aihua Liu, Xiao Liu, Juan Zhao, Yajun Xue, Bing Wei, Mingxia Zhang, Ru Gao, Hong Zhao, Jie Lu, Fan Li, Yang Zhang, Yiming Wang, Lei Zhang, Fengwei Tian, Jie Hu, Xin Gou contributed equally to this work.

Extended author information available on the last page of the article.

For the example in Zhang et al. (2021), there are 25 diseases causing nasal obstruction (chief complaint). In Table 9 in Zhang et al. (2021), 4 common diseases (chronic nasosinusitis, chronic rhinosinusitis with nasal polyps, allergic rhinitis and chronic hypertrophic rhinitis) proportion 98.5% of the total 3,214 case records of the 25 diseases. If we test the diagnostic precision of a medical AI system by randomly selecting cases from the 3,214 case records or we test all the 3,214 cases, 98.5% of the tested cases are the 4 common diseases, which means that if the diagnoses for the 4 common diseases are correct, the total diagnostic precision can be 98.5%, even though the diagnoses for other 21 uncommon diseases are incorrect. Obviously, this is not what we need, because the diagnostic precision in terms of diseases is only 4/25. It is hard for DNN to have high diagnostic precisions for uncommon diseases, because DNN has to overcome the problem of overfitting.

Moreover, "external validation" mentioned in Rajpurkar et al. (2022) is important, because medical AI should be applied in various scenarios, from large hospitals to village clinics. It should be validated that a medical AI can be applied in different scenarios with different data dimensions corresponding to different medical checks. In other words, invariance/generalization of medical AI in different scenarios is necessary for real applications. It is noted that DNN is based on the independent and identically distributed (i.i.d.) data assumption (Schölkopf et al. 2021). However, different scenarios may not satisfy the i.i.d. assumption. How to ensure the invariance/generalization of a medical AI is a serious challenge. In our understanding, the third-party (external) verification and real-world applications are necessary to justify the invariance/generalization. The best solution may be that the medical AI has the inherent invariance in different scenarios, just like a clinical expert who can diagnose diseases in different scenarios with his/her invariant professional knowledge without i.i.d. problem. With this invariance, we can verify the medical AI in high dimension cases (e.g., the retrospective verification with the discharged patient case records of the highest-level hospitals (the grade IIIA hospitals in China)) and ensure by algorithm that the verified medical AI is applicable in lower dimension cases (e.g., the cases of primary hospitals/clinics).

Causality-driven approach is promising to solve problems of "interpretability, transferability, robustness, and fairness" (Li et al. 2023). One of the reasons is that causality is usually invariant in different application scenarios and can perform counterfactual inference. Refs. (Schölkopf et al. 2021) and (Li et al. 2023) review many progresses in this research area. But only causal discovery models based on machine learning are addressed. Why do not we use the existing professional medical knowledge/causalities to construct a medical AI model, instead we extract causalities from data? It is seen that the traditional rule-based expert system has a lot of problems, such as fragmentation of knowledge representation, lack of rigorous algorithms for uncertainty propagation, lack of overall mathematical model, inefficiency in inference, etc. However, these do not mean that we should give up the use of expert's professional knowledge/causalities. Note that causal discovery faces a lot of problems such as data quality, high dimensions, causal complexity, large scale, etc.

To overcome the above problems and provide a trustworthy medical AI for clinical diagnosis, DUCG was developed (Zhang et al. 2021; Zhang 2012, Zhang et al. 2014, Nie and Zhang 2021, Dong et al. 2014, Zhang 2015a, b, Hao et al. 2017, Zhang and Yao 2018, Zhang et al. 2018, Dong and Zhang 2020, Qiu and Zhang 2021, Jiao et al. 2020, Ning et al. 2020, Deng and Zhang 2020, Zhang and Jiao 2022, Bu et al. 2023a, b), verified by third-party hospitals and applied in real-world.

Another problem that a practical medical AI must face is how to obtain medical information/evidences for an individual patient step by step in the diagnosis process, or how to dynamically perform medical checks accurately for an individual patient. The intuitive way is to check the symptoms, signs, laboratory and image examinations for the most suspected disease or the most dangerous possible disease in the current stage, which is the ordinary thinking of human doctors. DUCG provides another way: Calculate the overall contribution of a potential medical check whose result either validates or invalidates possible diseases, considered the danger degree of each possible disease and the cost (including injury to patient) to do the medical check. Then, rank the calculated recommendation degrees for all potential medical checks, so that clinicians can choose from them. The recommendation algorithm of DUCG is presented in this paper.

Section 2 introduces the DUCG approach briefly. Section 3 presents the DUCG algorithm to recommend potential medical checks. Section 4 describes the method for the third-party verification on the diagnostic precisions of DUCG. Section 5 provides application results of DUCG in the real-world in China. Section 6 extracts the key idea of DUCG and outlines the future work.

## 2. Brief Introduction to DUCG

DUCG is resulted from diagnosing faults in nuclear power plants (NPP) to avoid accidents such as Three Mile Island Accident (Zhang et al. 1991), where spurious sensor signals may exist. DUCG is required to have the ability to diagnose novel faults never occurred before. This requirement is the same as for operators of NPP. No data-driven approach can be applied, because NPPs are high reliable and every plant is different from others, which means rare or unavailable fault data. Once a fault occurs, operators are required to diagnose the fault based on their knowledge about this NPP. The knowledge is mainly the causalities with uncertainties among various variables/signals such as flow rate, temperature, pressure, water level, valve state, etc. Based on the success of DUCG in fault diagnoses (Zhang et al. 2014, Dong and Zhang 2020, Zhao et al. 2014, Zhang and Geng 2015, Zhang and Zhang 2016, Zhao et al. 2016, Zhao et al. 2017, Zhou and Zhang 2017, Dong et al. 2018, Han et al. 2023, Dong and Zhou 2023), DUCG is extended to diagnose diseases.

The basic model of DUCG is briefly illustrated in Fig. 1. In which, $V_{ij_i}$ is a parent event (parent variable $V_i$ in its state $j_i$); $X_{nk}$ is a child event (child variable $X_n$ in its state $k$); $F_{nk;ij_i} \equiv (r_{n;i}/r_n)A_{nk;ij_i}$ ; $A_{nk;ij_i}$ is the virtual independent causality event; $0<r_{n;i}\leq 1$ is the causal relationship intensity between $V_i$ and $X_n$; $r_n \equiv \sum_i r_{n;i}$ ; $X_{nk;ij_i}$ is a virtual event that $X_{nk}$ is just caused by $V_{ij_i}$ ; $a_{nk;ij_i} \equiv \Pr\{A_{nk;ij_i}\}$ ; $a_{nk;ij_i}$ and $r_{n;i}$ can be given by domain experts or learned from statistics (Zhang et al. 2018; Qiu and Zhang 2021). $V\in\{B, D, X, G, BX, SX, RG\}$. The DUCG variables and corresponding graphical symbols are described in Table 1. More details can be found in Zhang et al. (2021); Zhang et al. 2014; Dong et al. 2014; Zhang 2015a and (Deng and Zhang 2020).

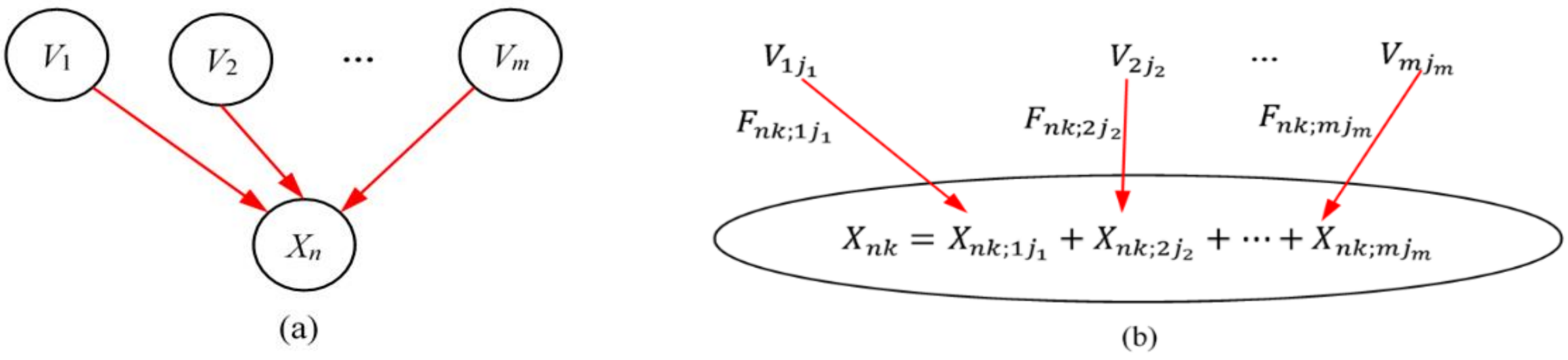

**Fig. 1** The basic mathematical model of DUCG, in which (b) describes the details in (a), $V$ represents parent variable/event, $X$ represents child variable/event, and $F$ represents the virtual functional variable/event between parent and child.

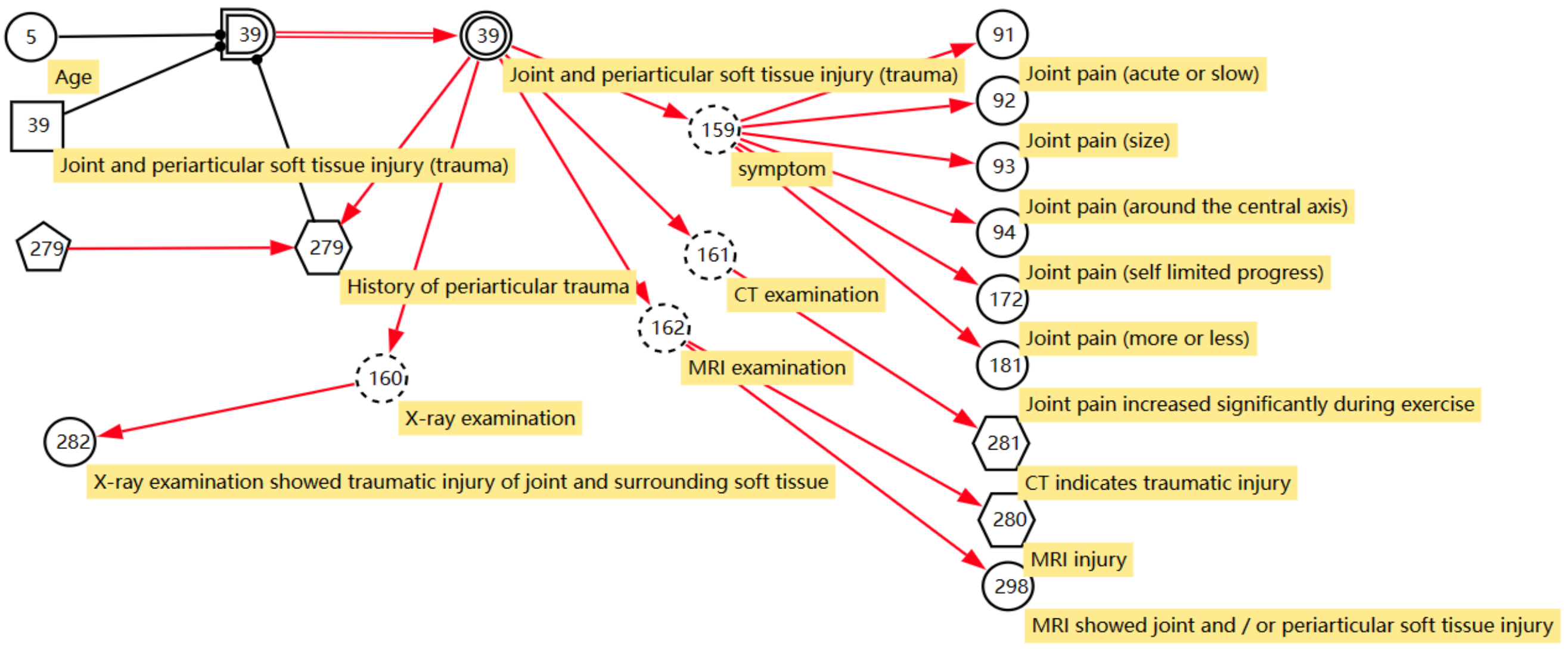

**Fig. 2** Illustrative example of a single-disease module/subgraph under the chief complaint arthralgia.

**Table 1** DUCG variables/symbols

| Symbol | Variable | Description |
|---|---|---|
| $n$ | $B_n$ | Basic/root variable/event representing disease |
| $n$ | $X_n$ | Effect variable/event, can also be cause variable/event |
| $n$ | $C_n$ | Virtual variable for classification |
| $n$ | $BX_n$ | The disease $B_n$ influenced by risk factors such as age, gender, medical histry, etc., so that its incidence is changed |
| $n$ | $G_n$ | Logic gate variable/event with at least two inputs, the logic relatonship is expressed in logic gate specification $LGS_n$ encoded in $G_n$ |
| $n$ | $SG_n$ | Specifial logic gate used to represent the combination of risk factors |
| $n$ | $D_n$ | Defaut/unknow cause variable/event |
| $n$ | $D_n$ | The added $D_n$ in inference when $X_{nk}$, $k$≠0, exists but no cause can be found |
| $n$ | $SX_n$ | Special $X$-type variable/event indicates a disease-specific manifestation |
| $n$ | $RG_n$ | Reversal logic gate with at least two outputs indicating concurent evidences |
| | $F_{n;i}$ | Functional event martix with $F_{nk;ij}$ as its member between cause/parent $i$ and effect/child $n$ |
| | $F_{n;i}$ | Conditional $F_{n;i}$ with $Z_{n;i}$ as the condition event |
| | $SF_{n;i}$ | Special $F_{n;i}$ , its function is to zoom in or zoom out Pr{$B_{ij}$} to be Pr{$BX_{ij}$} |
| | $SF_{n;i}$ | Condational $SF_{n;i}$ with $Z_{n;i}$ as the condition event |
| | | Input of $G$- or $SG$-type variable/event |

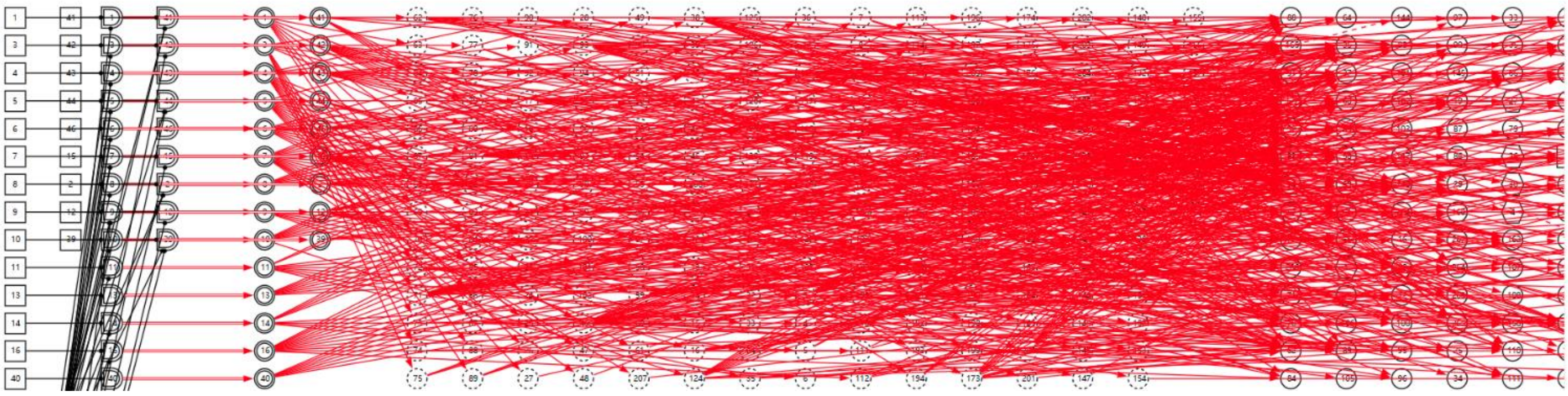

**Fig. 3** Part of a DUCG model synthesized by fusing same variables in different single-disease modules under a chief complaint.

With the symbols/variables shown in Table 1, we can separately and freely construct and update the modules for single-diseases under a chief complaint, and then synthesize them as a DUCG model of a chief complaint by fusing the same variables in different modules under the same chief complaint. An example of single-disease module/subgraph is shown in Fig. 2. A synthesized DUCG model for a chief complaint is shown in Fig. 3. Note that all single-disease modules are transparent and explainable. They are easy to be validated or invalidated by medical professionals. Then, the synthesized DUCG model under a chief complaint is also transparent and explainable. In other words, DUCG represents the understanding of human experts to the real world.

The construction and updating are implemented by clinical experts collaborating with DUCG technicians without data learning. The *a*-type, *r*-type and other type parameters are encoded in single-disease modules. Theoretically, these parameters can be learned from data as shown in (Zhang et al. 2018; Qiu and Zhang 2021). Practically, they are given by domain experts. Only the relative values of these parameters are meaningful, because the inference algorithm of DUCG is mainly in the form of numerator divided by denominator. In general, domain experts are good in giving the relative values but not the absolute values. Where, precise values are not needed, because the probability ranking of the diagnosed possible diseases is more important than the accurate probability values.

As illustrated in Fig. 1, child event $X_{nk}$ can be expanded as in Eq. (1):

$$X_{nk} = \sum_i \sum_{j_i} F_{nk;ij_i} V_{ij_i} = \sum_i \sum_{j_i} \left( r_{n;i} / r_n \right) A_{nk;ij_i} V_{ij_i} \tag{1}$$

More complex logical relationship among parent events are treated as a logic gate event $G_{ij_i}$ that is a virtual parent event of $X_{nk}$ as described in Table 1. The expanding of Eq. (1) can continue until $V \in \{B, D\}$.

In DUCG, the upper-case letters denote events/variables, and the lower-case letters denote probabilities of the corresponding events. For example, the probability form of Eq. (1) is Eq. (2):

$$x_{nk} \equiv \Pr\{X_{nk}\} = \sum_i \sum_{j_i} \left( r_{n;i} / r_n \right) a_{nk;ij_i} v_{ij_i} \tag{2}$$

In which,

$$\begin{aligned} a_{nk;ij_i} &\equiv \Pr\{A_{nk;ij_i}\} \\ v_{ij_i} &\equiv \Pr\{V_{ij_i}\} \end{aligned} \tag{3}$$

Note that $V \in \{B, D, X, G, BX, SX, RG\}$ and $v \in \{b, d, x, g, bx, sx, rg\}$, in which the *b*-type probability is the unconditional probability of a disease under a chief complaint and can be obtained from statistics, the *d*-type probability is defined as 1, because it is for the default event. Other probabilities can be calculated from Eq. (2) by replacing $x$ with $v$.

In principle, as shown in Eq. (4), the diagnosis of DUCG is to calculate the posterior probability $h_{kj}^s$ of hypothesis disease $H_{kj}$ (usually, $H$=$B$), conditional on evidence $E$:

$$\begin{aligned} h_{kj}^s &\equiv \Pr\{H_{kj} \mid E\} = \frac{\Pr\{H_{kj}E\}}{\Pr\{E\}} \\ E &= \prod_i E_i = \prod_i X_{ij_i} \end{aligned} \tag{4}$$

In which, $E_i = X_{ij_i}$ is a piece of evidence observed. The method to expand $E$ and $H_{kj}E$, which are a set of Eq. (1) multiplied together, is given in Zhang (2012) and Zhang et al. (2014), and is ignored in this paper. A DUCG recursive algorithm (Nie and Zhang 2021) can increase the computation efficiency of Eq. (4) greatly.

Before applying Eq. (4), the simplification to the DUCG of a chief complaint model should be done. To illustrate the simplification, consider the DUCG shown in Fig. 4 (a), in which $E$=$X_{3,0}X_{4,1}X_{8,1}$ is observed. Fig. 4 (b) describes the detailed causalities between $X_3$ and $X_1$, $X_3$ and $X_2$, and $X_4$ and $X_3$. As described in Fig. 4 (b), we have Eqs. (5)-(10), in which "–" indicates "null" or "0".

$$A_{3;1} = \begin{pmatrix} A_{3,0;1,0} & A_{3,0;1,1} & A_{3,0;1,2} \\ A_{3,1;1,0} & A_{3,1;1,1} & A_{3,1;1,2} \\ A_{3,2;1,0} & A_{3,2;1,1} & A_{3,2;1,2} \\ A_{3,3;1,0} & A_{3,3;1,1} & A_{3,3;1,2} \end{pmatrix} = \begin{pmatrix} - & - & - \\ - & A_{3,1;1,1} & - \\ - & A_{3,2;1,1} & A_{3,2;1,2} \\ - & - & - \end{pmatrix} \tag{5}$$

$$a_{3;1} = \begin{pmatrix} a_{3,0;1,0} & a_{3,0;1,1} & a_{3,0;1,2} \\ a_{3,1;1,0} & a_{3,1;1,1} & a_{3,1;1,2} \\ a_{3,2;1,0} & a_{3,2;1,1} & a_{3,2;1,2} \\ a_{3,3;1,0} & a_{3,3;1,1} & a_{3,3;1,2} \end{pmatrix} = \begin{pmatrix} - & - & - \\ - & a_{3,1;1,1} & - \\ - & a_{3,2;1,1} & a_{3,2;1,2} \\ - & - & - \end{pmatrix} \tag{6}$$

$$A_{3;2}=\begin{pmatrix} A_{3,0;2,0} & A_{3,0;2,1} \\ A_{3,1;2,0} & A_{3,1;2,1} \\ A_{3,2;2,0} & A_{3,2;2,1} \\ A_{3,3;2,0} & A_{3,3;2,1} \end{pmatrix}=\begin{pmatrix} - & - \\ - & - \\ - & - \\ - & A_{3,3;2,1} \end{pmatrix} \tag{7}$$

$$a_{3;2}=\begin{pmatrix} a_{3,0;2,0} & a_{3,0;2,1} \\ a_{3,1;2,0} & a_{3,1;2,1} \\ a_{3,2;2,0} & a_{3,2;2,1} \\ a_{3,3;2,0} & a_{3,3;2,1} \end{pmatrix}=\begin{pmatrix} - & - \\ - & - \\ - & - \\ - & a_{3,3;2,1} \end{pmatrix} \tag{8}$$

$$A_{4;3}=\begin{pmatrix} A_{4,0;3,0} & A_{4,0;3,1} & A_{4,0;3,2} & A_{4,0;3,3} \\ A_{4,1;3,0} & A_{4,1;3,1} & A_{4,1;3,2} & A_{4,1;3,3} \\ A_{4,2;3,0} & A_{4,2;3,1} & A_{4,2;3,2} & A_{4,2;3,3} \end{pmatrix} =\begin{pmatrix} - & - & - & - \\ - & - & A_{4,1;3,2} & - \\ - & - & - & A_{4,2;3,3} \end{pmatrix} \tag{9}$$

$$a_{4;3}=\begin{pmatrix} a_{4,0;3,0} & a_{4,0;3,1} & a_{4,0;3,2} & a_{4,0;3,3} \\ a_{4,1;3,0} & a_{4,1;3,1} & a_{4,1;3,2} & a_{4,1;3,3} \\ a_{4,2;3,0} & a_{4,2;3,1} & a_{4,2;3,2} & a_{4,2;3,3} \end{pmatrix} =\begin{pmatrix} - & - & - & - \\ - & - & a_{4,1;3,2} & - \\ - & - & - & a_{4,2;3,3} \end{pmatrix} \tag{10}$$

Given $E=X_{3,0}X_{4,1}X_{8,1}$, the DUCG in Fig. 4 (a) is simplified as Fig. 4 (c), because, as given in Fig. 4 (b), $X_1$ and $X_2$ are not the parent variable of $X_{3,0}$, $X_{3,0}$ is not the parent of $X_{4,1}$, and $X_1$, $X_2$ and $X_{3,0}$ are eliminated. Then, $B_7$ is eliminated because it does not connect to the positive evidences $X_{4,1}X_{8,1}$. In other words, possible diseases are reduced from $\{B_5, B_6, B_7\}$ in Fig. 4 (a) to $\{B_5, B_6\}$ in Fig. 4 (c). The appendix in Zhang et al. (2021) lists 11 rules to simplify DUCG given $E$. Readers can find more simplification situations according to the 11 rules. Usually, state 0 indicates negative/normal, which is the observed state of most variables and does not have any causal input and output. Thus, in most cases, the $A$-type or the corresponding $a$-type matrices are sparse as shown in Eqs. (5)-(10), which means that the simplified DUCG can be much smaller and simpler than the original DUCG.

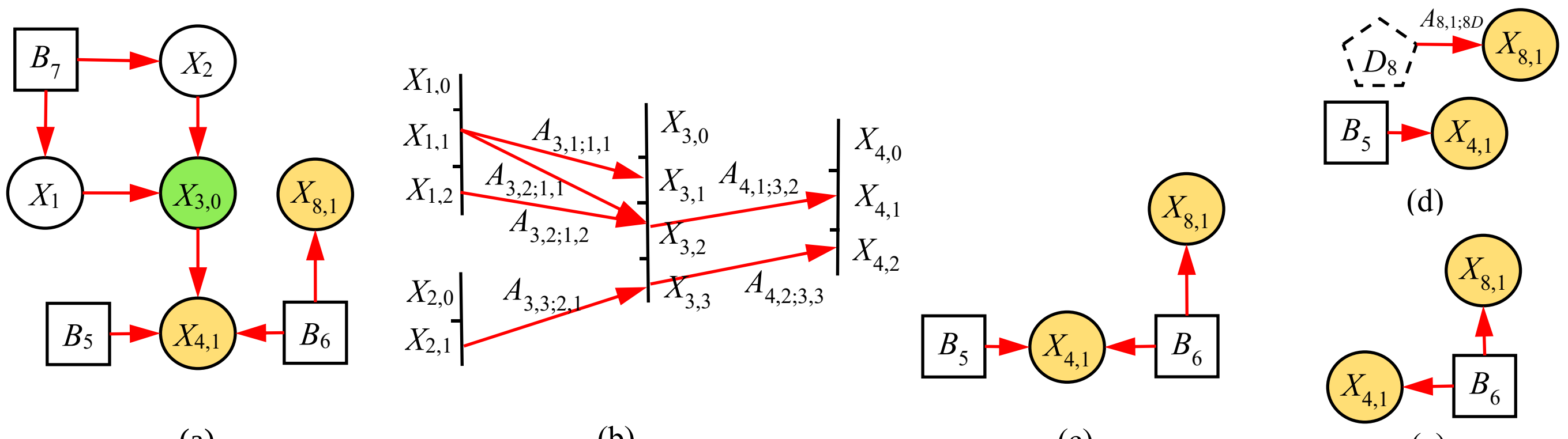

**Fig. 4** An illustration for DUCG simplification and separation given evidence $E=X_{3,0}X_{4,1}X_{8,1}$, in which (b) describes the detailed causalities connected to $X_3$ in (a), (c) is the simplified DUCG, (d) and (e) are two sub-DUCGs separated from (c) by assuming diseases $B_5$ and $B_6$ respectively, where green indicates negative state 0 and brown indicates positive state 1.

According to the one disease in one case assumption that is commonly used in clinical diagnoses as a principal, Fig. 4 (c) is further separated as two sub-DUCGs as shown in Fig. 4 (d) and (e). In the separation, the simplification rules are further applied. Note that in Fig. 4 (d), the positive/abnormal evidence $X_{8,1}$ is not caused by the assumed disease $B_5$ and is isolated. A virtual $D$-type event, i.e. $D_8$ along with $A_{8,1;8D}$ in Fig. 4 (d), is added as the cause of the isolated positive/abnormal evidence $X_{8,1}$ according to Rule 10[2], which reduces the suspicion degree of the assumed disease significantly (see Zhang et al. 2021 for details). The final sub-DUCGs explore all possible diseases conditional on the evidence $E$, provide explanations to these possible diseases, and are used to calculate the suspicion degrees of these possible diseases.

A realistic example of sub-DUCG is shown in Fig. 5, in which the state indices of variables are ignored. The green color nodes indicate the observed negative/normal states of variables. They were expected to be positive/abnormal with certain probabilities for the assumed disease and will decrease the suspicion degree of the assumed disease. The other color nodes indicate the observed positive/abnormal states of variables, which are as expected as in DUCG with certain probabilities for the assumed disease and will increase the suspicion degree of the assumed disease. In the left lower corner in Fig. 5, there are 5 isolated positive/abnormal evidences that cannot be caused by the assumed disease, which means that this disease may be much less possible.

[2] Rule 10: If $E$ shows $X_{nk}$ is true while $X_{nk}$ does not have any input due to any reason, add a virtual parent event $D_n$ to $X_{nk}$ with $a_{nk;nD}=1$ and $a_{nk';nD}=0$, $k\neq k'$. $r_{n;D}$ can be any value. The added virtual $D_n$ can be drawn as 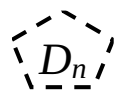 in the simplified graph.

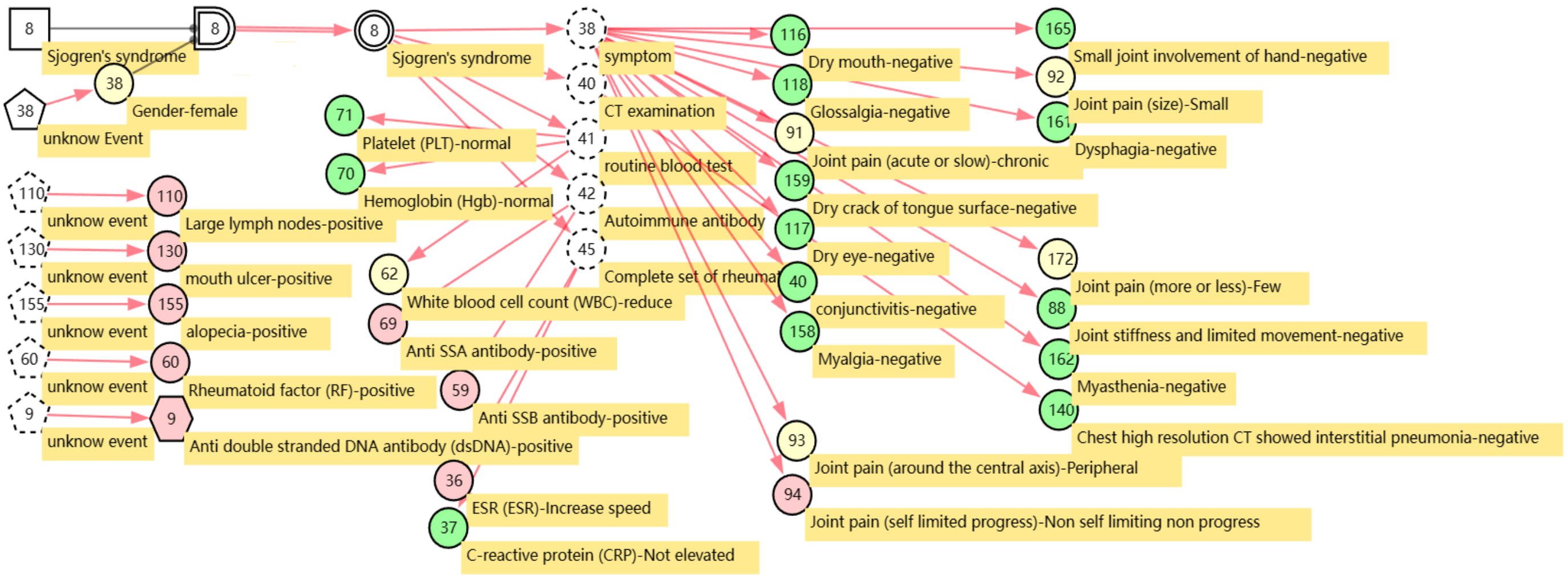

**Fig. 5** Example of an explainable diagnosis result that is a sub-DUCG by assuming the disease "Sjogren's syndrome" given $E$ shown as the color nodes in the sub-DUCG, in which the green nodes indicate negative/normal states, and the other color nodes indicate positive/abnormal states.

Index the current diagnosis step as $y$, $y$=1, 2, …, the suspicion degree $h_{kj}^{p}(y)$ of possible disease $H_{kj}$ can be calculated from Eq. (9) that is deduced from Eq. (4) (see Zhang et al. (2021) for details).

$$h_{kj}^{p}(y)=\varphi(y)\frac{\Pr\{E(y)\,|\,\text{sub-DUCG}_{kj}\}}{\sum_{H_{kj}\in S_H(y)}\Pr\{E(y)\,|\,\text{sub-DUCG}_{kj}\}} \qquad (9)$$
$$\varphi(y)=\sum_{i\in S_{XK}}\varepsilon_i\Big/\sum_{i\in S_{XK}+S_{XU}}\varepsilon_i$$

In Eq. (9), sub-DUCG$_{kj}$ indicates the sub-DUCG by assuming possible disease $H_{kj}$, $S_H(y)$ is the set including all possible hypotheses/diseases in step $y$, $\varepsilon_i$ is the attention degree of $X_i$, $SX_i$ or $RG_i$; $S_{XK}$ is the index set of state-known $X$- and $SX$-type variables; and $S_{XU}$ is the index set of the state-unknown $X$- and $SX$-type variables, and $\varphi(y)$ is the check completeness in step $y$. Details can be found in Zhang et al. (2021). Note that in Zhang et al. (2021), $h_{kj}^{p}(y)$ was improperly denoted as $h_{kj}^{s}(y)$ that is confusing with $h_{kj}^{s}$ in Eq. (4).

According to the suspicion degrees calculated from (9), we can rank all possible diseases in $S_H(y)$. The ranking along with suspicion degrees and explanations (sub-DUCGs) of possible diseases are the final diagnosed results.

It is seen that all parameters and calculations have clear physical meanings. This enables medical professionals to understand the diagnosed results, and validate or invalidate the knowledge representation and inference algorithm of DUCG. Once an incorrect diagnosis is found, we can trace the diagnosis process and check single-disease modules to find what the mistake is. After corrections, we can ensure that the same incorrect diagnosis will no longer occur.

In Fig. 4, suppose $E=X_{1,0}X_{2,0}X_{3,0}X_{4,1}X_{8,1}$. Figs. 4 (d) and (e) can still be obtained by applying the simplification rules and separating diseases $B_5$ and $B_6$, because $X_{1,0}$ and $X_{2,0}$ (negative/normal states of $X_1$ and $X_2$) do not have input and output like $X_{3,0}$. In this new case, two new evidences $X_{1,0}$ and $X_{2,0}$ are added. Compared to the early $E=X_{3,0}X_{4,1}X_{8,1}$, the new case has 5 dimensional observations, and the early case has 3 dimensional observations. That is, for a same DUCG, no matter how many dimensional evidences can be observed, we can use the same causalities represented in the DUCG to make diagnosis, just like a clinical expert to diagnose diseases in different scenarios with his/her invariant knowledge. The DUCG should be verified in high dimensions, so that the causalities can be verified as more as possible. Then, we can apply these causalities to diagnose diseases in different scenarios with same or reduced dimensional observations/evidences.

It is easy to understand that the knowledge/causalities represented in DUCG are invariant in different application scenarios. If they are variant, we need to construct different DUCGs for different scenarios. For example, some diseases are in south but not in north and vice versa. Then we need to construct the south version DUCG and north version DUCG. Fortunately, most signal-disease modules are the same in both south and north.

Since the DUCG construction does not need to collect, process and label, and learn from huge amount of case records and other data, the cost and time of DUCG constructions are reduced dramatically. Compared to the data-driven approaches, DUCG's hardware requirement and energy consumption are ignorable. The most expensive part of the whole work is the cost for DUCG technicians including software engineers and clinical experts. DUCG needs high level clinical experts, because they determine the upper limit of DUCG.

A shortage of DUCG is that DUCG cannot recognize medical images and sounds. The current solution is to provide referential images, sounds and videos for users to refer to and compare with. Uncertain evidences are allowed in DUCG. The DUCG algorithm to deal with uncertain evidences is presented in Zhang 2015b. In the future, DUCG can collaborate with data-driven approaches to assist clinicians to recognize medical images and sounds, thus to complete the whole process of intelligent diagnoses.

Since the single-disease modules are constructed in a same way for common and uncommon diseases, there is no problem that diagnostic precision for uncommon disease is less than the common disease in principle. The only problems are: (1) the verification case records for uncommon diseases are less than the common diseases, resulting in that the uncommon diseases are less and even not verified; and (2) the lack of knowledge for diagnosing uncommon diseases.

Finally, how to obtain $E(y{=}y{+}1)=E(y)E^{+}(y)$ step by step is what we need to discuss in the following section, where $E^{+}(y)$ denotes the next observed evidences.

## 3. Algorithm to Recommend Potential Medical Checks

The recommendation algorithm of DUCG is presented in Eq. (10):

$$I_i(y)=\frac{\beta_i\rho_i(y)}{\sum_{i\in S_X(y)}\beta_i\rho_i(y)} \tag{10}$$

$$\rho_i(y)=\frac{1}{\lambda_i(y)}\sum_{H_{kj}\in S_H(y)}\omega_{kj}\sum_{g\in S_{iG}(y)}\Pr\{X_{ig}\mid E(y)\}\left|h_{kj}^{p}(X_{ig}E(y))-h_{kj}^{p}(E(y))\right|$$

In Eq. (10), $y$ indexes the current stage, $I_i(y)$ is the recommendation degree for the potential medical check to the state-unknown $X_i$ or $SX_i$, $S_X(y)$ is the index set of state-unknown $X_i$ and $SX_i$; $\beta_i$ scores the cost (including the injury to patient) to do the medical check for observing the state of $X_i$ or $SX_i$; $S_{iG}(y)$ is the index set of $g$ in $X_{ig}$ that is a possible state of $X_i$ as a result of the medical check; $\omega_{kj}$ is the danger degree of disease $H_{kj}$; $S_H(y)$ is the set of possible $H_{kj}$. $\lambda_i(y)$ is the number of possible diseases that may cause $X_i$. $\beta_i$ and $\omega_{kj}$ are given by clinical experts. $\beta_i$ can be changed on demand of an individual patient. The rank of $I_i(y)$ guides the accurate medical checks to diagnose disease step by step. According to Eq. (10), $E^{+}(y)$ can be obtained.

The physical meaning of $\rho_i(y)$ in Eq. (10) is: Suppose we check state-unknown variable $X_i$. The probability that the check result is $X_{ig}$ is $\Pr\{X_{ig}|E(y)\}$. Add $X_{ig}$ into $E(y)$ so that the new evidence is $E(y{=}y{+}1)=X_{ig}E(y)$. Calculate the absolute difference between the suspicion degree with new evidence, i.e. $h_{kj}^{p}(X_{ig}E(y))$, and the suspicion degree without new evidence, i.e. $h_{kj}^{p}(E(y))$. This difference has included the information of the absolute value of $h_{kj}^{p}(E(y))$. More difference means more value to check $X_i$. Weight the difference by $\Pr\{X_{ig}|E(y)\}$. Sum up the weighted difference for all states of $X_i$. Multiply the weighted difference with the danger degree of disease $H_{kj}$, i.e. $\omega_{kj}$. Sum up the results for all possible diseases indexed by $H_{kj}\in S_H(y)$. Divide the sum by $\lambda_i(y)$, which means that the more possible diseases connecting to $X_i$, the less value to check $X_i$, because the more the check result validates or invalidates connected diseases, the less the check result tells us about which disease is more possible. Then $\rho_i(y)$ is calculated. It is obvious that the more the weighted difference is, the more $\rho_i(y)$ is, and the more dangerous $H_{kj}$ is, the more $\rho_i(y)$ is. Therefore, $\rho_i(y)$ represents the value to check $X_i$. Note that $\rho_i(y)$ has considered all possible diseases included in $S_H(y)$. Therefore, $\rho_i(y)$ is more comprehensive than considering only the dangerous and high possible diseases.

Finally, $I_i(y)$ is the recommendation degree to check $X_i$, in which the value $\rho_i(y)$ and the cost $\beta_i$ to check $X_i$ are considered. Users can select which state-unknown variables to check according to the rank of the recommendation degrees and local condition.

No example is provided in this paper to illustrate the recommendation algorithm, because the calculation is too complex. The mathematical and physical meanings of Eq. (10) are clear enough for readers to understand.

Now, we can summarize the whole diagnosis process of DUCG as shown in Fig. 6, in which the update of DUCG is implemented by human experts instead of by adding data into a machine. In this way, we know what, where and why to make changes to the DUCG system, and can evaluate the influence of the changes.

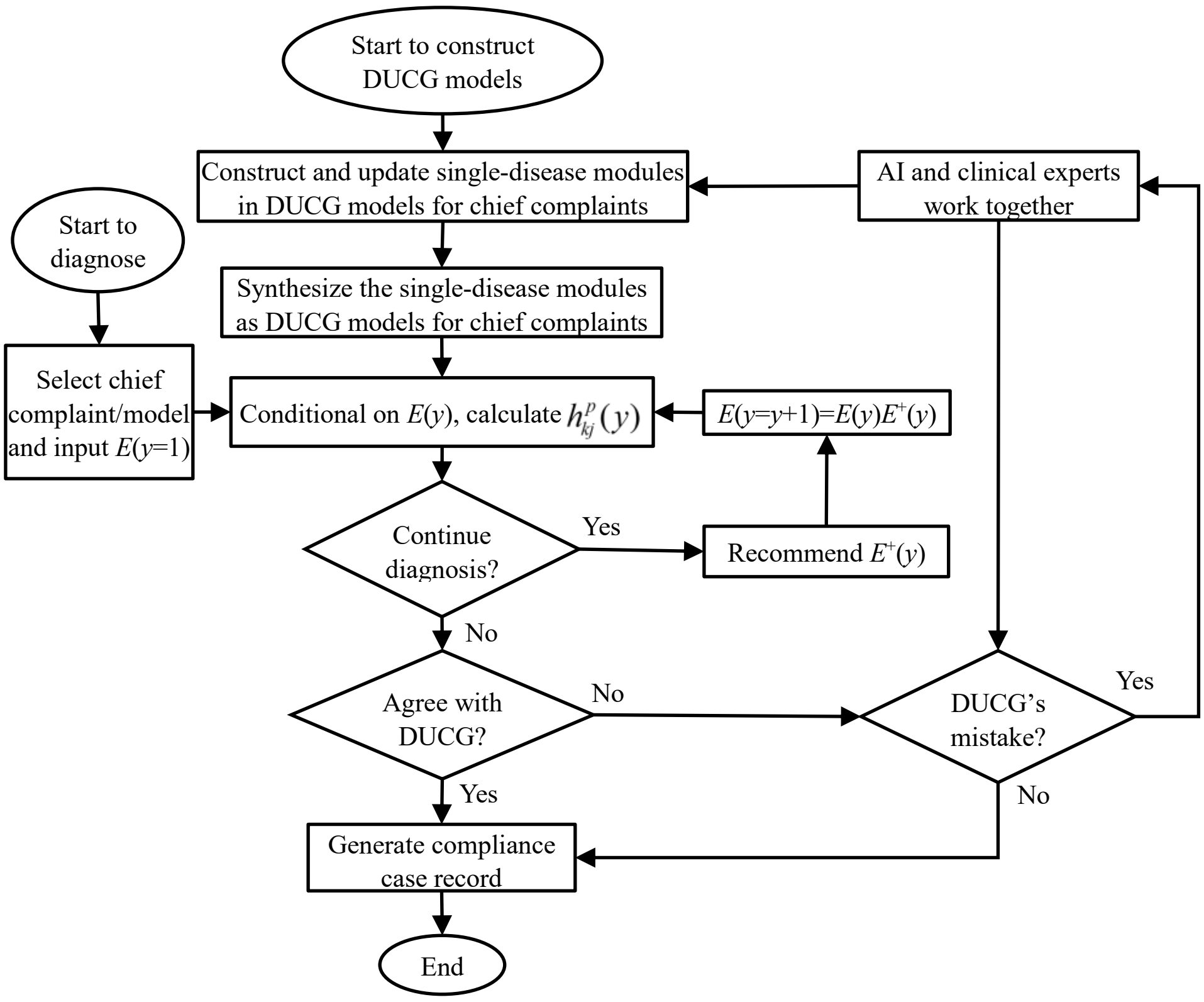

**Fig. 6** The flow chart of DUCG diagnosis

# 4. Third-Party Verifications on Diagnostic Precisions

As mentioned in Rajpurkar et al. (2022), the external or third-party verifications for the diagnostic precisions of the medical AI system is very important, because it can help to find the influence of the i.i.d. assumption and data bias. The so-called third-party means the independent hospitals who have nothing to do with the construction of the DUCG models and whose data are not used in the DUCG system. To ensure the quality of the verification, the third-party hospital should be in the highest grade, i.e. grade IIIA in China. Only the discharged case records should be used for the verification, not the outpatient case records, because the latter's quality is uncertain.

The method for the third-party to verify the diagnostic precisions of a DUCG model is as follows:

(1) Search case records in the Electronic Medical Record (EMR) system with the chief complaint as the same as one of the chief complaints of the DUCG model being verified (sometimes, a group of related chief complaints are included in one DUCG model);
(2) Sort the searched case records out according to the diseases included in the DUCG model;
(3) Randomly select 10 qualified case records of a disease in the DUCG model for the testing ("qualified" means that the information recorded supports the diagnosis). If the number of searched qualified case records is less than 10, all searched qualified case records are selected. If no qualified case record is searched out, give up the verification for this disease;
(4) Manually input the information recorded in the selected case record into DUCG, and check the diagnosis result of DUCG to see if the diagnosed disease raking first is the same as in the case record. If yes, this case is accounted for correct. If not, analyze the diagnosis result of DUCG by the clinical experts of the third-party to see if the DUCG's result is correct. If yes, this case is accounted for correct. Otherwise is accounted for incorrect;
(5) Calculate the diagnostic precision for every tested disease according to the number of the cases accounted for correct divided by the total number of tested cases for the disease;
(6) Calculate the diagnostic precision of the DUCG model: Sum up the number of the tested cases accounted for correct for all diseases in the DUCG model. Divide the sum by the number of total tested cases no matter they are accounted for correct or not;
(7) Certify the results and stamp the verification report by the third-party hospital.

According to the method above, we have verified 46 DUCG models covering 54 chief complaints covering more than 1,000 diseases covering more than 10,000 ICD-10 (International Classification of Diseases version 10) disease codes. The results are: the diagnostic precisions of all the 46 DUCG models are no less than 95%, in which the precision for every disease including uncommon one is no less than 80%.

It is very important to test every disease as equal number of case records as possible, instead of randomly selecting case records from EMR systems, because common diseases make up the majority. If we randomly select case records without number limitation for a disease (the limitation in this paper is 10), the verified precision can be high, even though all diagnoses for uncommon diseases are incorrect or not selected. Since the uncommon diseases are rare, higher limitation cannot increase the number of tested cases for uncommon diseases and can only increase the number of tested cases for common diseases.

In this paper, "uncommon" is conditional on the chief complaint and may become "common" under other chief complaint.

An example of the third-party verification for the chief complaint nasal obstruction is reported in Zhang et al. (2021). There are 25 diseases that may cause nasal obstruction in this DUCG model. Table 9 in [8] shows that 4 out of 25 diseases (nasopharyngeal angiofibroma, fracture of frontal sinus, fracture of ethmoidal sin, atrophic rhinitis) do not have qualified case records in the EMR system of the third-party hospital. 88 case records for the other 21 diseases are searched out and tested, in which only one was incorrect. Thus, the diagnostic precision of this DUCG model is 87/88=98.86%. Meanwhile, except the four diseases without qualified case records, 20 diseases have 100% diagnostic precision and 1 disease (acute sinusitis) has 80% diagnostic precision.

Similarly, 13 DUCG models (arthralgia, dyspnea, cough and expectoration, epistaxis, rash, abdominal pain, hematochezia, diarrhea, nausea and vomiting, chest pain, sore throat, fever, palpitations) are verified by seven grade IIIA hospitals organized by Chongqing Science and Technology Bureau under two research projects (Chongqing is a direct city of China). These hospitals are all independent of the DUCG construction that is done in Beijing far away from Chongqing. 424 diseases are included in the 13 DUCG models, in which 77 diseases did not have qualified case records searched out from the EMR systems. The diagnostic precisions of all the tested diseases are 100%.

It is reasonable for DUCG to have 100% diagnostic precisions, because DUCG is transparent and modularized. Once an incorrect diagnosis is identified, the mistake in DUCG can be found and corrected. Here we need to emphasize that the so-called correct means to be consistent with the clinical experts' judgement. DUCG does not guarantee the absolute correctness.

The above verifications are only retrospective. No prospective study has been completed, because of the limited budget, time and conditions. We will do the prospective studies in the future researches. However, the feedback from hundreds of clinicians who apply DUCG in the real-world for more than one million cases compensates the absence of prospective studies to some extent. As shown in Fig.7, there is a mechanism to receive feedback from users/clinicians. Once they disagree with the diagnosis of DUCG, they are encouraged to report the case to us (the action is just to click a button on the screen) and we will discuss the case with the clinician and analyze the case to see whether the DUCG diagnosis is incorrect. If yes, the mistake will be found and corrected. In fact, 17 incorrect diagnoses have been identified. All of them were traced, and the mistakes in DUCG were found and corrected. After the corrections, no same incorrect diagnosis has been reported. This will be addressed in the next section.

# 5. Real-World Applications

There are 46 DUCG models that have been constructed under chief complaints, verified by third-parties, and then applied in the real-world in China. They are:

Cough sputum, dyspnea, abdominal pain, diarrhea, hematemesis, nasal congestion, nasal bleeding, blood in the stool, nausea and vomiting, joint pain, hemoptysis, fever, chest pain, jaundice, anemia, edema, obesity, emaciation, sore throat, palpitation, fever in children, dizziness, headache, constipation, rash, difficulty swallowing, enlargement of lymph nodes, cyanosis, limb numbness, vaginal bleeding, abnormal vaginal discharge, pruritus vulvae, reduced menstruation or amenorrhea, abdominal distension, syncopation, tinnitus, deafness, earache, acid reflux, heartburn, hiccup, belching, mass, oliguria or no uria, lower urinary tract symptoms (frequent urination, urgency of urination, pain in urine, dysuria, polyuria, gross hematuria, and urine leakage), neck and low back pain (neck pain, waist pain and back pain).

In which, 44 models include single chief complaint respectively, and 2 models include a group of related chief complaints respectively. In total, 54 chief complaints are included. Each DUCG model includes 20+ to 100+

diseases that are across hospital divisions and may cause a same chief complaint. After removing duplicates, more than 1,000 diseases are included, covering more than 10,000 ICD-10 disease codes.

Since 2020, the 46 DUCG models have been gradually, after third-party verifications respectively, applied in the real-world in Jiaozhou of Qingdao in Shandong, Zhongxian of Chongqing, and other areas in China, covering hundreds of village clinics, grade I hospitals and grade II hospitals. These types of medical units take more than 70% of total diagnoses in China. By the end of 2023, 1.06 million real diagnosis cases with DUCG were performed. Only 17 were identified as incorrect, in which 12 were the cases that the DUCG model did not include the corresponding diseases at that time, e.g. pelvic inflammation was not included in the abdominal pain model; 4 were the incorrect causalities leading to the incorrect diagnoses; 1 was a misassigned disease code of ICD-10. These mistakes were found and corrected. After the corrections, no further same incorrect diagnosis cases have been reported.

In Jiaozhou, by the end of 2023, the number of diagnosis cases with DUCG were more than 660,000. In which, the disagreement ratio was 0.05%. The local clinicians were encouraged to report the disagreement cases. In the reported disagreement cases, 54 were incorrect application of DUCG models, e.g. applying arthralgia model for headache, because the headache model was not applicable at that time; 80 were chronic diseases that did not need diagnosis; 23 were incorrect information input, e.g. some default selections of negative states of variables should be positive; 191 were mistaken as incorrect but were finally confirmed as correct through discussions with us; 7 were confirmed as incorrect, and the mistakes were found and corrected. In terms of the number of diagnosis cases, the top 20 DUCG models in Jiaozhou are shown in Fig. 7.

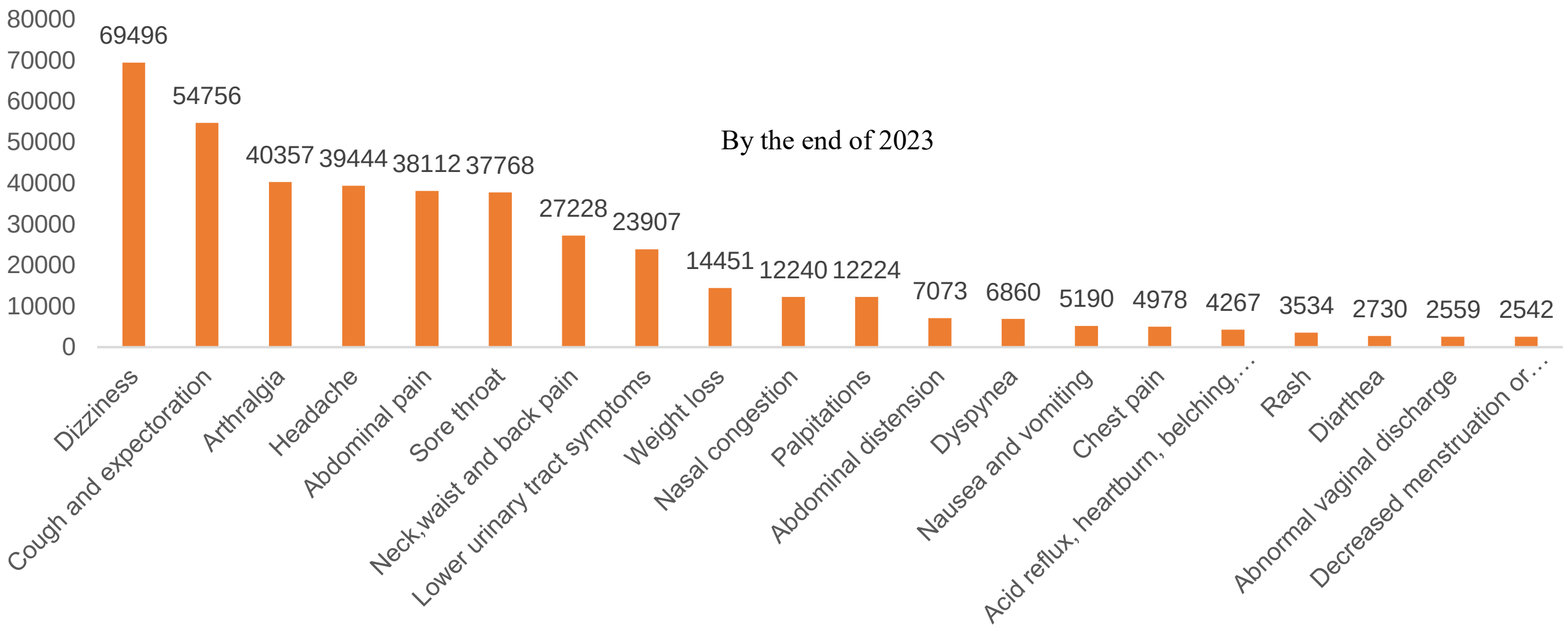

**Fig. 7** The top 20 DUCG models used in Jiaozhou according to the number of application cases.

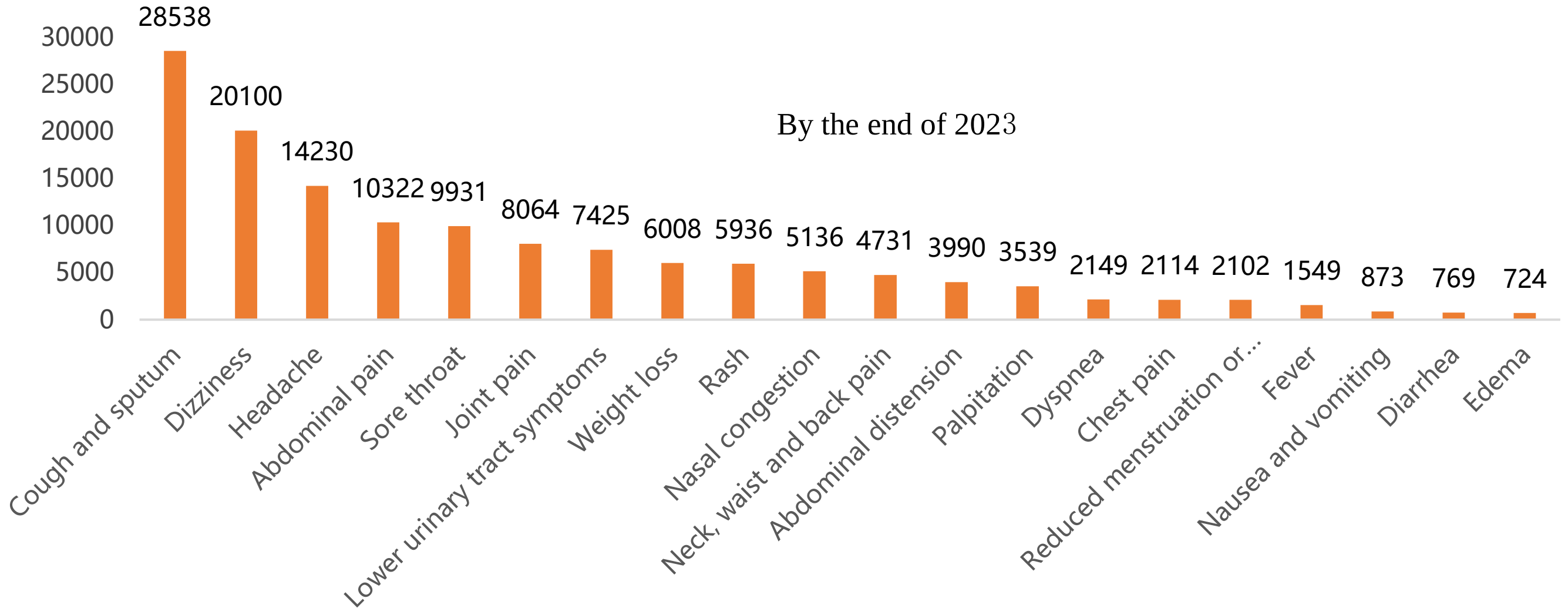

**Fig. 8** The top 20 DUCG models used in Zhongxian according to the number of application cases.

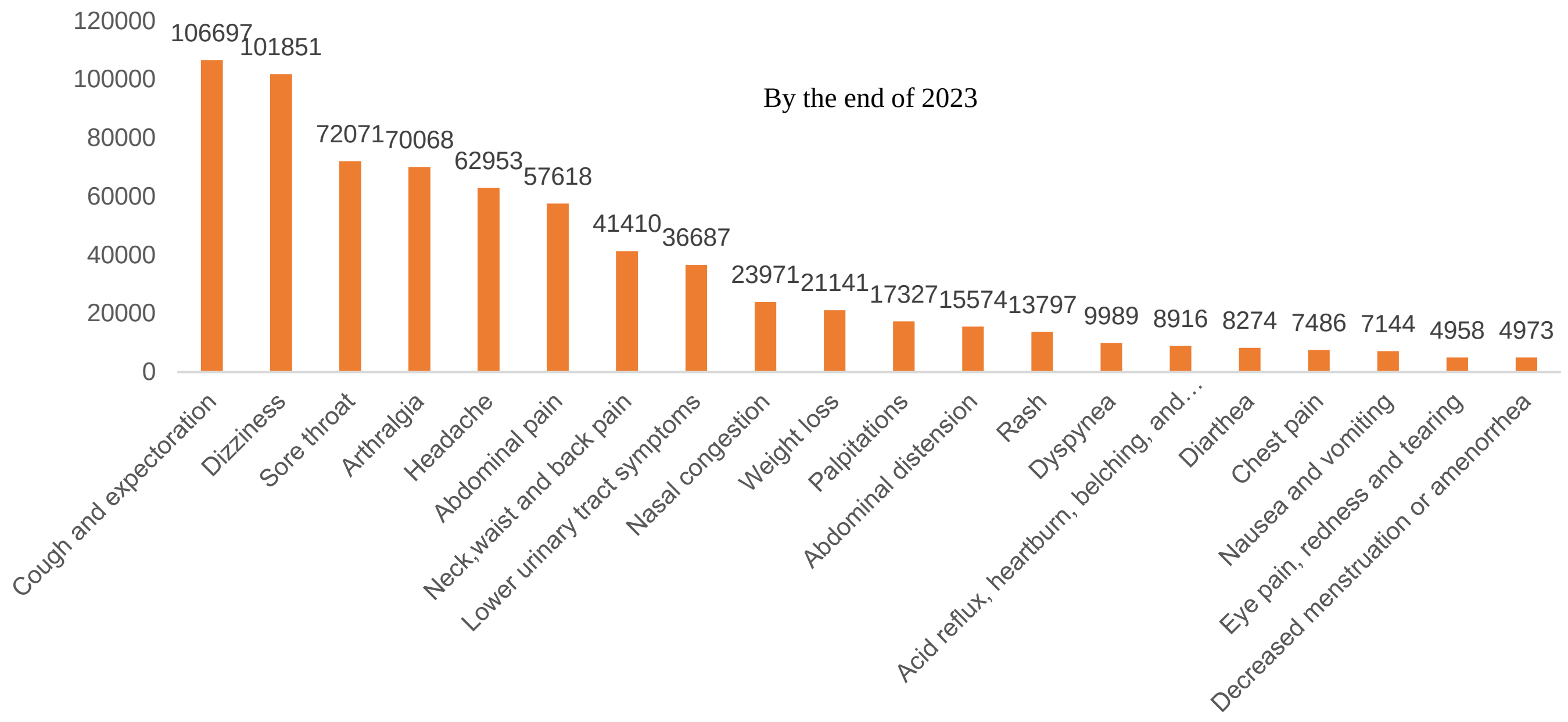

**Fig. 9** The top 20 DUCG models used in total according to the number of application cases.

In Zhongxian, by the end of 2023, the number of diagnostic cases with DUCG was more than 156,000. There were no village-level applications because the medical internet was unavailable for village clinics. The disagreement ratio was 0.15%. In the reported disagreement cases, 45 were incorrectly selecting DUCG models; 14 were chronic diseases without need for diagnosis; 18 were incorrect information input; 151 were mistaken as incorrect; 7 were confirmed as incorrect and the mistakes were corrected. The top 20 DUCG models in terms of the number of diagnosis cases in Zhongxian are shown in Fig. 8.

Finally, the top 20 DUCG models in terms of the number of diagnosis cases of all areas are shown in Fig. 9.

**Definition:** Define ability improvement rate (IR) as the number of diseases diagnosed by applying DUCG in a year divided by the number of diseases diagnosed without DUCG in 2019, minus 1.

Table 2 shows the IRs of the clinicians who applied DUCG in 2021 and 2022 respectively, where clinicians who applied DUCG refer to those who applied DUCG to diagnose disease at least once. In a same year, a clinician who applied DUCG might also diagnose diseases without applying DUCG. As a result, the total number of diseases diagnosed by clinicians who applied DUCG at least once might be more than shown in Table 2, which means that the IRs might be higher than in Table 2 if we consider the diagnosed diseases without applying DUCG.

Chronic diseases, including high blood pressure, coronary heart disease and diabetes, were excluded from the calculation for IR, because they had usually been known when patients went to see clinicians and no diagnosis was needed. The purpose of these patients to see clinicians is to take medicine.

The distributions of the average IR of clinicians in different ranges in terms of the number of cases applying DUCG are shown in Fig. 9 for Jiaozhou and Fig. 10 for Zhongxian respectively. “Average” means the sum of IRs of clinicians in a case number range of applying DUCG divided by the number of clinicians in that number range.

The IR was negative for the clinicians who applied DUCG within a few hundred cases in a year. This was because when they applied DUCG within a few hundred cases, the number of diseases diagnosed by applying DUCG was unlikely to be more than that in much more diagnosis cases in 2019 without DUCG. Some clinicians might apply DUCG only when they were unconfident in diagnosing diseases.

The IR does not decrease via the increased case number of applying DUCG. The reason may be that the number of diseases DUCG can diagnose is much more than a clinician can diagnose without DUCG. In fact, DUCG can diagnose more than 1,000 diseases, while most clinicians without DUCG can diagnose less than 100 diseases. Table 3 shows 6 selected examples of clinicians applying DUCG, in which HIS means hospital information system. Note that not all diseases diagnosed by DUCG were recorded in HIS.

**Table 2** Ability improvement rate (IR) in Jiaozhou and Zhongxian in 2021 and 2022 respectively

| Areas | Number of clinicians who applied DUCG | Number of diseases diagnosed by the same group of clinicians in 2019 without DUCG | Number of diseases diagnosed by the same group of clinicians who applied DUCG | IR |
|---|---|---|---|---|
| Jiaozhou | 2021: 223 | 244 | 472 | 93.44% |
| | 2022: 253 | 240 | 589 | 145.4% |
| Zhongxian | 2021:172 | 265 | 473 | 64.91% |
| | 2022: 85 | 233 | 286 | 22.75% |

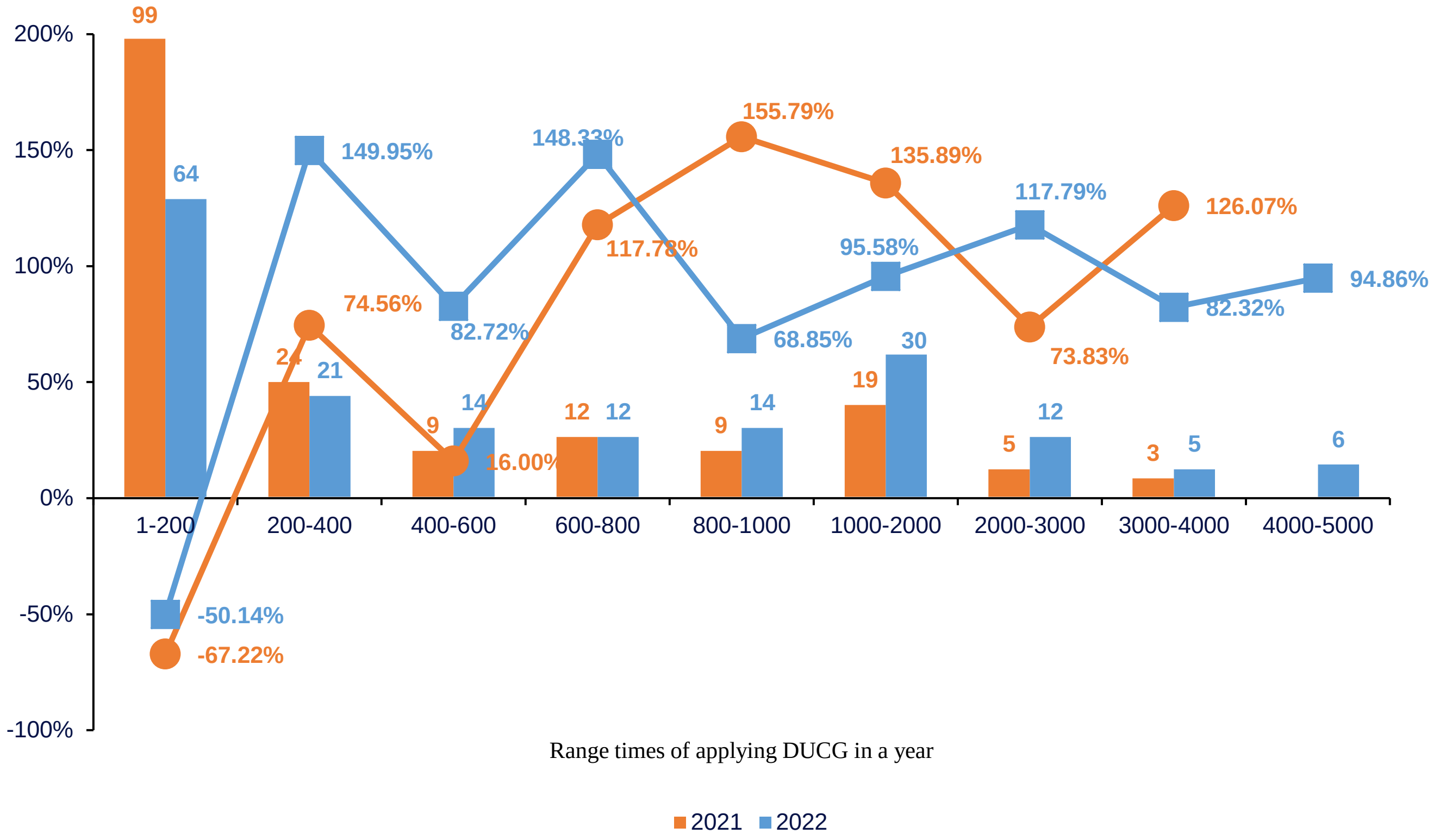

**Fig. 10** The numbers and average IR distributions of local clinicians in different ranges of applying DUCG in Jiaozhou in 2021 and 2022 respectively.

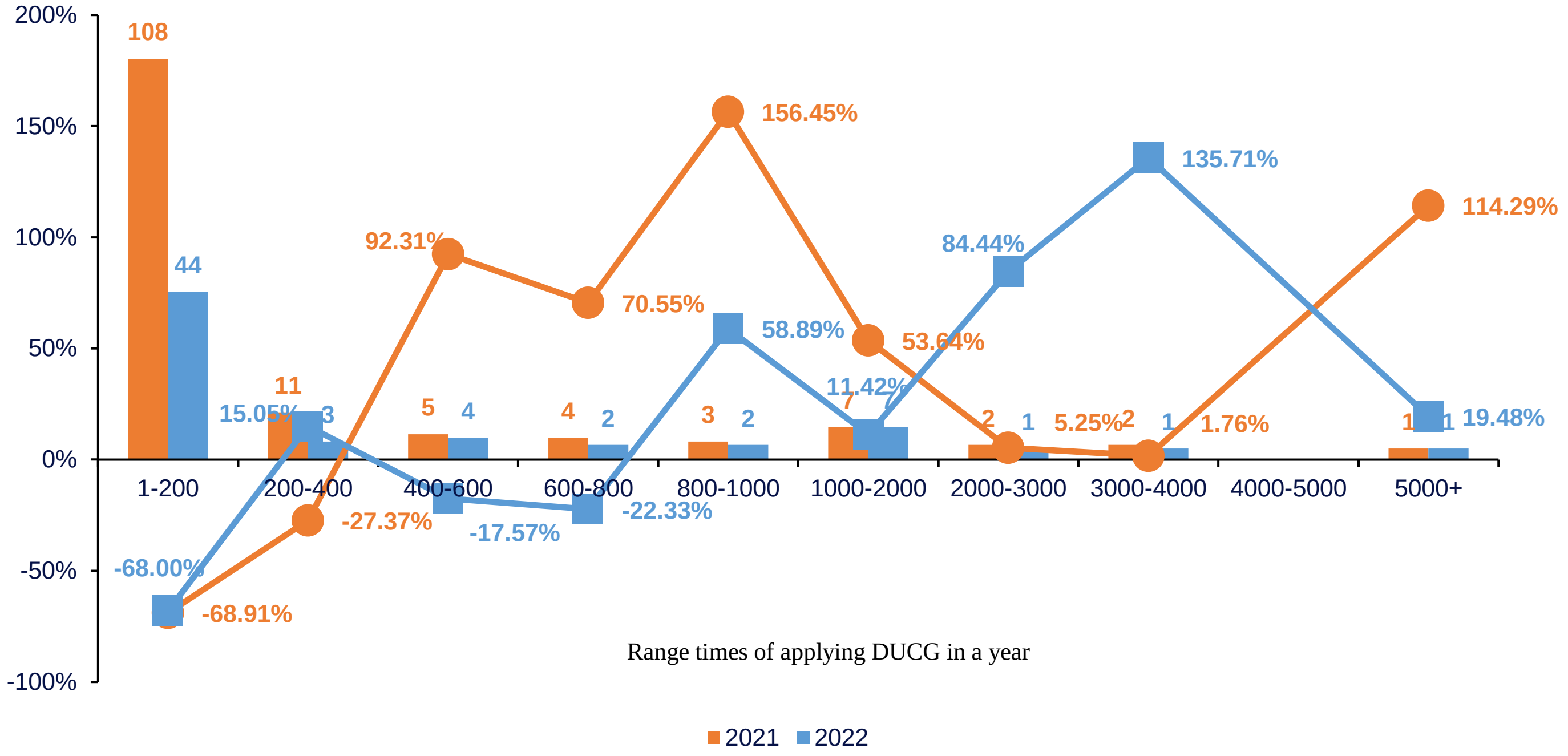

**Fig. 11** The numbers and average IR distributions of local clinicians in different ranges of applying DUCG in Zhongxian in 2021 and 2022 respectively.

**Table 3** Selected local clinicians who applied DUCG

| Area | Clinician | Number of diagnosed cases in 2021 and 2022: DUCG (chronic diseases) \| HIS (chronic diseases) | Without DUCG in 2019 | Applying DUCG in 2021 | | Applying DUCG in 2022 | |
|---|---|---|---|---|---|---|---|
| | | | Number of diagnosed diseases | Number of diagnosed diseases | IR | Number of diagnosed diseases | IR |
| Jiaozhou | Zhao | 2021: 2736 (1267) \| 3733(1780)<br>2022: 3935 (2311) \| 4465(2636) | 64 | 92 | 43.8% | 102 | 59.4% |
| | Peng | 2021: 1713 (485) \| 2820(845) | 24 | 87 | 263% | 90 | 275% |
| | | 2022: 1462 (574) \| 1421(563) | | | | | |
| | Dai | 2021: 1468 (639) \| 1755(957)<br>2022: 1730 (1319) \| 1475(1096) | 18 | 67 | 272% | 51 | 183% |
| Zhongxian | Zhou | 2021: 5217 (1054) \| 5993(1824)<br>2022: 5674 (1399) \| 5608(1740) | 77 | 165 | 114% | 92 | 19.5% |
| | Tan | 2021: 1399 (7) \| 1372(26)<br>2022: 2007 (31) \| 2152(244) | 45 | 88 | 95.6% | 83 | 84.4% |
| | He | 2021: 842 (635) \| 848(632)<br>2022: 567 (443) \| 557(428) | 19 | 65 | 242% | 31 | 63.2% |

The highest IR was from village clinician Ma in Jiaozhou. He diagnosed 12 diseases without DUCG in 2019, and 88 diseases by applying DUCG in 2021. The IR was: 88÷12−1= 633%. Because of COVID19, his data was incomplete in 2022.

The applications in Jiaozhou was better than in Zhongxian. The reason might be (1) there were no village clinics who applied DUCG in Zhongxian, because they were unable to connect to DUCG through medical internet; (2) we had less time to train Zhongxian's clinicians to use DUCG, because the transportation is difficult and the influence of COVID19 was more serious in Zhongxian.

The data of 2023 are under analyses. It is difficult to compare the number of diseases diagnosed by applying DUCG with the number of diseases diagnosed without DUCG, because we need to classify the diseases diagnosed without DUCG as the diseases in DUCG, except that some (if any) of them are not included in DUCG. It is found that the text descriptions for the diseases diagnosed without DUCG are very chaotic.

Finally, the hardware requirement (a sever) to run DUCG costs less than $10,000, which can fulfil applications and concurrent demand for a county area (e.g. Jiaozhou or Zhongxian where population is up to 700,000). The computation is efficient (within 1s per diagnosis).

## 6. Key Idea and Future Work

The unique key idea of DUCG is to represent and deal with uncertain causalities at the basic layer rather than at the appearance layer, thus to decouple complex correlations among variables and parameters. Due to the decoupling, the modularized construction for large and complex DUCG can be implemented, so do the simplification, separation, logic operation in inference and update in any module.

The so-called appearance layer is the statistical layer. For example, Bayesian network [38] and causal Bayesian network [39] use statistical conditional probability tables (CPTs) in the directed acyclic graph (DAG) to express the joint probability distribution (JPD) over variables. DNN is another form of the appearance layer model.

The so-called basic layer is the basic causal mechanism layer. DUCG introduces a virtual independent random causal functional event ($A$-type event) to represent the basic uncertain causal mechanism between a parent event and its child event. The occurrence probability of the $A$-type event, i.e. the $a$-type parameter, quantifies the uncertainty of the basic causality. $A$ and $a$ are local and independent of other variables and parameters. The combination of various independent events and their occurrence probabilities constitutes the JPD, CPTs, etc., and thus decouples variables and parameters coupled at the appearance layer.

The future work is planned as follows:

(1) Perform prospective studies to further verify the diagnostic precisions of DUCG;
(2) Apply DUCG in more areas and continue to improve it, including to do more third-party verifications;
(3) Collaborate with data-driven medical AI, so that the useful information included in medical images and sounds can be extracted as the input of DUCG;
(4) Develop more DUCG models for rare disease diagnoses as shown in Ning et al. (2020);
(5) Develop traditional Chinese medicine DUCG;
(6) Develop English and other language versions of DUCG (so far, the DUCG in applications is only in Chinese).

## Acknowledgements

This research was supported by Beijing Yutong Intelligence Technology Co., Ltd., Institute for Guo Qiang, Tsinghua University (project number: 2020GQG0001), Chongqing Science and Technology Bureau (project numbers: cstc2018jscx-mszdx0106 and cstc2019jscx-dxwtBX0018), National High Level Hospital Clinical Research Funding (2022-PUMCH-A-017).

**Author contributions** Z. Zhang presented the classification algorithm and other algorithms of DUCG, developed and conducted the development and implementation of the DUCG cloud platform, model constructions, third-party verifications and real-world applications; Q. Zhang originally presented the DUCG framework and conducted the whole project; The following authors mainly contributed to the constructions, verifications and applications of the models: Y. Jiao with abdominal pain, arthralgia, fever, neck and back pain, dizziness, diarrhea, limb numbness, large lymph nodes, syncope, oliguria, dyspnea; L. Lu with edema, obesity, emaciation; L. Ma with mass; A. Liu with headache; X. Liu with abdominal pain; J. Zhao with hematemesis, jaundice, rash, fever; Y. Xue with chest pain; B. Wei with cough sputum, hemoptysis, dyspnea, cyanosis; M. Zhang with nasal obstruction, nasal bleeding, sore throat, tinnitus, deafness, earache; R. Gao with diarrhea, bloody stool, nausea and vomiting, constipation, dysphagia, abdominal distention, acid reflux, heartburn, hiccup, belching; H. Zhao with fever, anemia; J. Lu with heart palpitation; F. Li with dizziness, syncope; Y. Zhang with vaginal bleeding, abnormal vaginal discharge, pruritus vulvae, reduced menstruation or amenorrhea; Y. Wang with frequent urination, urgent urination, urination pain, dysuria, polyuria, gross hematuria, urine leakage; L. Zhang with child fever. F. Tian, J. Hu, and X. Gou mainly contributed to the third-party verifications. The authors listed in 3–18 contribute equally to this work and are all the third authors of this paper.

## Authors and Affiliations

Zhan Zhang[1] · Qin Zhang[2] · Yang Jiao[3] · Lin Lu[4] · Lin Ma[5] · Aihua Liu[6] · Xiao Liu[7] · Juan Zhao[8] · Yajun Xue[9] · Bing Wei[10] · Mingxia Zhang[10] · Ru Gao[11] · Hong Zhao[10] · Jie Lu[12] · Fan Li[13] · Yang Zhang[14] · Yiming Wang[15] · Lei Zhang[16] · Fengwei Tian[17] · Jie Hu[18] · Xin Gou[19]

Correspondence author: Qin Zhang
qinzhang@tsinghua.edu.cn

1 Department of Computer Science and Technology, Tsinghua University, Beijing, China
2 Institute of Nuclear and New Energy Technology & Department of Computer Science and Technology, Tsinghua University, Beijing, China
3 Department of General Internal Medicine, Peking Union Medical College Hospital, Chinese Academy of Medical Sciences & Peking Union Medical College, Beijing, China
4 Department of Endocrinology, Key Laboratory of Endocrinology of National Health Commission, Union Medical College Hospital, Chinese Academy of Medical Sciences & Peking Union Medical College, Beijing, China
5 Peking Union Medical College Hospital, Chinese Academy of Medical Sciences & Peking Union Medical College, Beijing, China
6 Beijing Neurosurgical Institute, Beijing Tiantan Hospital, Capital Medical University, Beijing, China
7 Department of Gastroenterology, Beijing Hospital, National Center of Gerontology, Institute of Geriatric Medicine, Chinese Academy of Medical Science, Beijing, China
8 Beijing YouAn Hospital, Capital Medical University, Beijing, China
9 Department of Cardiology, Beijing Tsinghua Changgung Hospital, School of Clinical Medicine, Tsinghua University, Beijing, China
10 Xuan Wu Hospital of Capital Medical University, Beijing, China
11 Department of Gastroenterology, Beijing Chao-Yang Hospital, Capital Medical University, Chaoyang District, Beijing, China
12 Department of Special Medical Treatment Center, Fuwai Hospital, National Center for Cardiovascular Diseases, Chinese Academy of Medical Sciences and Peking Union Medical College, Beijing, China
13 Peking University First Hospital, Peking University, Beijing, China
14 Peking University People's Hospital, Peking University, Beijing, China
15 Department of Urology, China Rehabilitation Research Center, Beijing, China
16 Capital Institute of Pediatrics, Beijing, China
17 Chongqing Traditional Chinese Medicine Hospital, Chongqing, China
18 Suining Central Hospital, Suining, Sichuan, China
19 Department of Urology, The First Affiliated Hospital of Chongqing Medical University, Chongqing, China